\documentclass[conference, compsoc]{IEEEtran}
\usepackage{ifpdf}
\usepackage{cite}
\usepackage[pdftex]{graphicx}
\DeclareGraphicsExtensions{.pdf,.jpeg,.png}
\usepackage[cmex10]{amsmath}
\usepackage{algorithmic}
\usepackage{array}
\usepackage{mdwmath}
\usepackage{mdwtab}
\usepackage[tight,footnotesize]{subfigure}
\usepackage[font=footnotesize]{subfig}
\usepackage{fixltx2e}
\usepackage{stfloats}
\fnbelowfloat
\usepackage{url}

\begin{document}
\title{A Cognitive Mind-map Framework to Foster Trust}
\author{\IEEEauthorblockN{Jayanta Poray, Christoph Schommer}\IEEEauthorblockA{Dept. of Computer Science and Communication\\ÊILIAS Laboratory - MINE Research Group\\ University of Luxembourg\\L-1359 Luxembourg, Luxembourg}
}
\maketitle

\begin{abstract}
The explorative mind-map is a dynamic framework, that emerges automatically from the input, it gets. It is unlike a verificative modeling system where existing (human) thoughts are placed and connected together. In this regard, explorative mind-maps change their size continuously, being adaptive with connectionist cells inside; mind-maps process data input incrementally and offer lots of possibilities to interact with the user through an appropriate communication interface. With respect to a cognitive-motivated situation like a conversation between partners, mind-maps become interesting as they are able to process stimulating signals whenever they occur. If these signals are close to an own understanding of the world, then the conversational partner becomes automatically more trustful than if the signals do not or less match the own knowledge scheme. In this (position) paper, we therefore motivate explorative mind-maps as a cognitive engine and propose these as a decision support engine to foster trust.
\end{abstract}

\section{Explorative Mind-maps}
The principles of  explorative mind-maps $M_i$ have already been described in \cite{SB05}, where we accent that mind-maps rely on the natural principle on sensations and the corresponding propagation of stimuli to a final destination. Indeed, explorative mind-maps share this principle through an associative architecture that incrementally processes accepted stimuli to a consistent informational structure. This is similar to the natural paradigm, but on contrast to a verificative processing of a user's thoughts, the explorative mind-maps are built from the bottom up, meaning that their existence exclusively interdepend on incoming signals.

\begin{figure}[!h]
   \centering
   \includegraphics[width=8.5cm]{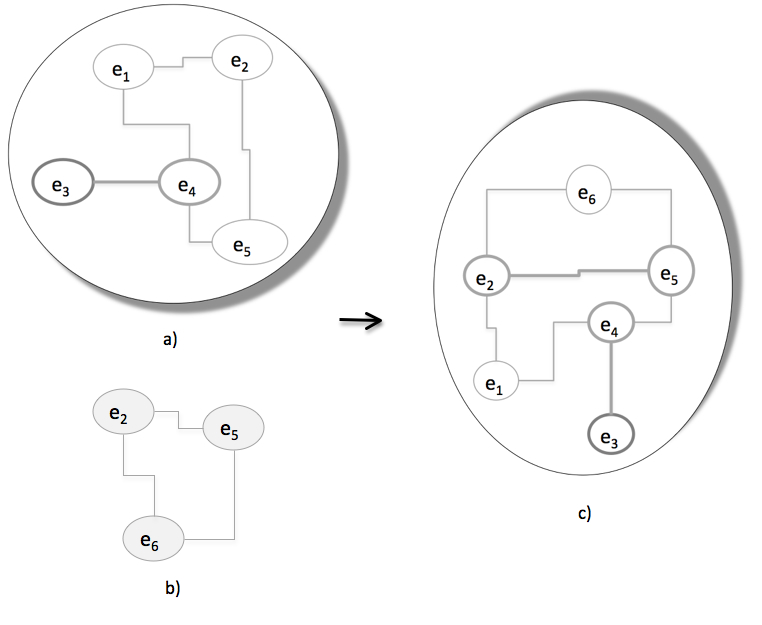} 
   \caption{Merge between the existing mind-map a) and a (newly) mini-network b) to an updated mind-map c). Entity cells $e_2$ and $e_5$ are higher activated; the connection in between has been learned, the activation increased as well.}
   \label{fig:img10}
\end{figure}

Explorative mind-maps share a sub-symbolic architecture that is composed of interacting entity cells $e_i$. As mentioned above for the natural principle, these cells foster on a processing of data streams and a stimulation/inhibition-principle of adjacent connections. The activation of such a connectionist architecture bases on a dynamic construction of cell structures during the processing of the input stream.

In the \emph{stimulation phase}, a stream data is stimulated and absorbed by \emph{receptor (input) cells} $r_i$, which decompose the stream to its entities. For example, the text streams are decomposed into the  word entities, transactional streams to item entities, and so on. Using \emph{filter cells} $f_i$, those receptor cells $r_i$ are inhibited that do not address a semantic interest. In the \emph{Mini-Network phase}, the collection of entities, which occur at such a specific time-point, form a mini-network \cite{KEM01} with fully connected mini-network cells $m_i$. The \emph{Mind-map Merger} starts once the mini-network is established: in this phase, the mini-network is sent to the mind-map and is merged with the existing entity cells in the mind-map (initially, the mind-map is empty). In this regard, the specified merge references an action of mini-network and entity cells ($m_i$ and $e_i$) - that share a same representation - to a unique entity cell. The activation status $act(e_i)$ of an entity cell is then increased, in case it has been merged. If two adjacent entity cells $e_k$ and $e_l$ of the mind-map are activated by the mini-network at the same time, their connection weight $\omega(e_k,e_l)$ is increased by a learn rate $\phi$ (Hebbian Learning). If two adjacent entity cells remain inactivated, their connection weight is decreased by a fragment of $\phi$. The association becomes \emph{forgotten} if the connection weight is below the forget parameter $\sigma$. The mind-map may degenerate due to less intensive stimuli.

The usage of \emph{memories} depends on skeletons $k_j(t)$, which are higher-activated clumps of entity cells that occur temporally and that consist of entity cells sharing a strong activity. As for the natural example, these skeletons may be sent to memories: depending on its state and temporal eligibility, the mind-map may keep existing skeletons inside the \emph{short-term memory (STM)} or send it to the \emph{long-term memory (LTM)}. Whereas the \emph{short-term memory} comprehends itself as a location on short-notice, the \emph{long-term memory} is a place of long-duration that is only accessible to skeleton patterns being permanently present or recurring. 

In a final phase, during communication with the outside world, comparison between the actual state of the mind and the outside contents is realised. This is to be done either on demand - the user explicitly sends a request - or unsolicited. In so far, the consequence of the brief descriptions above is not only that explorative mind-maps, which are \emph{non-deterministic} regarding its size, appearance, and communication. Also that explorative mind-maps stand for a life-cycle process that depends on the intensity of incoming stimulations and the activity inside the mind-map.   

\begin{figure}[!h]
   \centering
   \includegraphics[width=8cm]{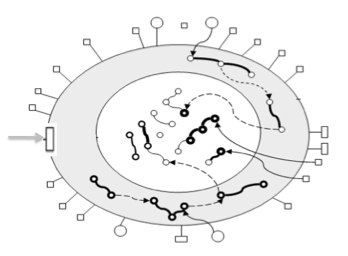} 
   \caption{The explorative Mind-maps with the \emph{STM} (grey area) and \emph{LTM} (inner area).}
   \label{fig:img7}
\end{figure}

Figure \ref{fig:img7} presents a figurative demonstration of the mind-map. When the corresponding mini-networks arrive on the left side, they are already assigned to the receptor cells $r_i$ and filtered to $f_i$ cells. These mini-networks are entered into the outer area \emph{STM}. Inside the \emph{STM}, mini-networks are compared with the individual mind-map structures one by one. Note that these mind-maps are temporarily taken from the \emph{LTM} (inner core) part. If the mini-network skeleton ($k_j(t)$) supports the structure of any existing mind-map, the skeleton is properly mapped and merged into the mind-map itself. Here, depending on the strength and relevance, clumps of the entity cells (skeletons $k_j(t)$) may be sent to the \emph{LTM} from the  \emph{STM} and/or kicked out to the outer area. Once being in the outer area, they (the skeleton or its subset) are released from the mind-map.

Currently, some implementations, which are described in \cite{BHSWW08}, have been realised and tested. First, an intrusion detection system has been designed and implemented. The explorative mind-map is used as an internal communication center for circulating immune cells in order to exchange novel intruder-related patterns and news. A second system manages implicit and explicit user feedback (inside a search retrieval environment) by a backbone mind-map engine. In a third system, an author-centric graph system - which is the mind-map - for bibliographic entries layer has been implemented to demonstrate affinity between social communities.

However, all these mind-map implementations allow an exploited sub-role within a larger software system, focusing on a given purpose. This is,  in our opinion - a non-optimal view, because mind-maps satisfy the demand for a cognitive engine that may work independently and non-deterministically.

\section{May we understand Explorative Mind-maps as an Engine of Trust? }

%
\subsection{Mind-maps meet the cognitive demands!}
Generally, cognition is commonly accepted as to be the process of thought. In the wide field of Artificial Intelligence, it refers to the representation of mental processes (thoughts) and functions that state intelligent entities.

The acceptability of explorative mind-maps with these matters is that they have already demonstrated cognitive abilities, they are adaptive to the input and work continuously and incrementally (like humans and creatures). Explorative mind-maps disarm the claim for a real-time system by a sliding window of size \emph{k} and the filter cells. It also address the complexity problem by a dynamic and flexible association affectation. In general, each processing step is done in parallel while having a synchronization step. Mind-maps are associative, dynamic, and fault-tolerant; they share a hybrid structure with a sub-symbolic mind-map core and an user-directed interaction, which is not only related to a symbolic parameter adjustment but also to a verificative retrieve of existing mind-map information. And, whenever a mini-network is sent to the mind-map, some of the existing entity cells change their activation status and their activation to the neighbor entity cells through learning.

Even though the existent explorative mind-maps are mostly deterministically organized (as the procedure of what to do is predetermined), we want to find an answer to the question on how to integrate advanced cognitive performances like for example trust. With trust, we understand the assumption that future changes will have a positive and expected progression. Furthermore, we understand trust as the result of an existing reliable, authentic and confident status, which follows from a representation of the current internal knowledge (belief), the state of personal emotion and associated interest. As mention in \cite{JB06} and \cite{CFP03}, the individual decision (to trust or not) is additionally governed by a host of factors like past experiences, history of events, others' opinion and socio-cognitive behavior.

\subsection{When \emph{Alice} converses with \emph{Bob}}
The trust between two persons is considered as the measure of a mutual belief. How to define the appropriate trust opinion during a conversation is a matter of concern.  Assume that two natural persons \emph{Alice} and \emph{Bob} talk to each other where both may decide if an incoming conversational signal is considered as worth to remember or even not. Also if textual patterns inside the conversational streams are to be extracted, summarized and kept in mind - or even handled as noise. If both \emph{Alice} and \emph{Bob} store conversational signals inside a knowledge-based representation framework, then \emph{Alice} will know something (or something more) about \emph{Bob} and vice versa. After receiving the input signal from \emph{Bob}, \emph{Alice} surely develops some certain belief about \emph{Bob} and his believes - at least with respect to the subject of conversation (and vice versa). Moreover, we may follow the idea that both are able to match their own mind-maps (representation of what each of them thinks) against the mind-maps of the conversing partner: both may decide at any time if they trust, and if yes, up to which level.

\section{A first approach}
To realize such mind-maps, however, we need a strategy to concern with the raw data input. To filter out stop-words and to morphologically reduce a word to its root seems to be manageable by the inclusion of a thesaurus or a dictionary. Also the concern with typical natural language problems like the occurrence of anaphors or the ambiguity of word senses - in both cases, we must find the corresponding entity - are problems of larger consequences. Presently, we pay no attention to it but set it back thereinafter. Earnestly, we presently ask for the calculation of the match while only figuring out an existing mind-map. We are aware that a trust function must take into account a temporal stamp, because the trust decision may be revised every time.

\subsection{Constituting two Mind-maps}
We claim that each person who converses must have (at least) two independent mind-maps $M_p$ and $M^{*}_{pq}$ that affect both an individual understanding about his/her own world ($M_p$) and some estimations about the world of the conversing partner ($M^{*}_{pq}$). Actually $M_p$ is a fixed wired system that includes own beliefs, emotions, etc. (firstly in a word-related representation that is built by words) and also $M^{*}_{pq}$ is the own representation of the conversing partner , strictly speaking, of the related words within a conversation.

With regard to this, a person's trust is obtained if the own representation of the conversing partner $M^{*}_{pq}$ is fully consistent with the own mind-map $M_p$. On the other side, a person may not agree to the conversing partner if the gap between the matching mind-maps get too wide.


In case that a person $p$ converses with two persons, namely $q$ and $r$, corresponding mind-maps (for the outer world) $M^{*}_{pq}$ and $M^{*}_{pr}$ will be created for maintain individual conversational view. 
Most of the real time conversational applications are supported by this model, where $p$ needs to converse with a group of persons at the same time. 

\subsection{\emph{gtrust}: A suggestive matching function}
In a naive approach, the decision of trusting another person may depend on the number of entity cells that is in $M_p$ and $M^{*}_{pq}$. Here two mind-maps are called similar, if final number of matching entity cells reaches certain threshold value. This degree of similarity exactly defines the trustworthiness of $p$ for the conversational partner $q$. Furthermore, an improved version of this naive approach, however, is to take into account the activation status (and the corresponding connection status). The similarity between the two mind-maps $M_p$ and $M^{*}_{pq}$ is then not only the pure number of common entity cells but moreover a weighted sum.

The advantage of such an approach is that all nodes of the mind-map are used with the identical privilege. The disadvantage is that keywords or key statements are not focused on adequately. Another suggestion therefore is to select only $k$ entity cells with the highest frequency - or the other way around - those entity cells with the lowest frequency (assuming that less occurring words are more interesting than others). Alternatively, we may sample the entity cells to select randomly selected nodes.

All strategies are worth to become applicable, but often a person typically owns a number of entities that are highly important whereas other entities - although being important - remain negligible. An example hereby is the color of a car, which is in many cases one of the KO criterions (to a woman) but even less important than the amount of horsepowers (to a man). We therefore appeal to arrange an individual \emph{parallel universe}, which consists of entity cells with corresponding \emph{self-relevance} $\rho^{M_p}(e_i,t)$ for its own entity cell $e_i$ and the \emph{outer-relevance}  $\rho^{M^{*}_{pq}}(e_j,t)$ for the entity cell $e_j$, achieved from the conversational partner $q$. Each relevance value is dynamic and may change over time $t$, being strong or weak: the strongest entity cells then resemble to neural attractors, the weakest cells to ``fellow-runners''.

With respect to a simulative engine in terms of a cognitive mind-map, the mentioned parallel universe does not exist in the beginning of the mind-map life-cycle. Consequently, any relevance values of any entity cell is not initially given. Undoubtedly, the change of dynamic value is to be handled in internally, whereas the initialization step must be examined explicitly.

So far, several alternatives have been identified, which do either use an a-priori knowledge or demand an ex-posterior information. A first (and in our opinion the most appropriate) approach is to randomize each relevance value, which has the advantage that a-priori less/no efforts must be investigated. The belief that more important entity cells adapt to a higher relevance value (and less interesting entity cells to lower relevance values) can not be dismissed. The idea, to initially differ between entity cells explicitly is a tedious concern. Also it does not correspond to the evolution of mind-maps per se. An ex-posterior initialization of relevance values, for example by guessing or communicating with the conversing partner, may be feasible but is less applicable. Then, we propose the function $match(M_p, M^{*}_{pq}, t)$ as follows:

\begin{equation}
match(M_p, M^{*}_{pq}, t) = \sum_{i=1}^{k}\sum_{j=1}^{l} \frac{|\rho^{M_p}(e_i,t) \cap  \rho^{M^{*}_{pq}}(e_j,t)|}{|\rho^{M_p}(e_i,t)|}
\end{equation}
Here $M_p$ is the self mind-maps for the person $p$ and $M^{*}_{pq}$ is the outer mind-maps for the person $p$, developed by the person $q$. Where $k$ and $l$ are the number of entity cells that contain the corresponding relevance values $\rho^{M_p}(e_i,t)$ and $\rho^{M^{*}_{pq}}(e_j,t)$ during the time slice $t$. The trust function is then to be evaluated as

\begin{equation}
 gtrust(M_p, M^{*}_{pq}, t)  = 
      \left\{ \begin{array}{ll} 
          yes, & if \mbox{ } match(M_p, M^{*}_{pq}, t) \geq \alpha \\
          no & else \end{array} \right.
\end{equation}

Please note that the function $match$ is time varying and the trust threshold $\alpha$ is individual and given to each conversational partner solely. In this context the trust threshold  $\alpha$ is a determinable parameter, which can be  influenced by several factors : for instance by  the number of entity cells, present inside the mind-maps. 

\subsection{An example}

\begin{figure}[!h] 
   \centering
   \includegraphics[width=5.5cm]{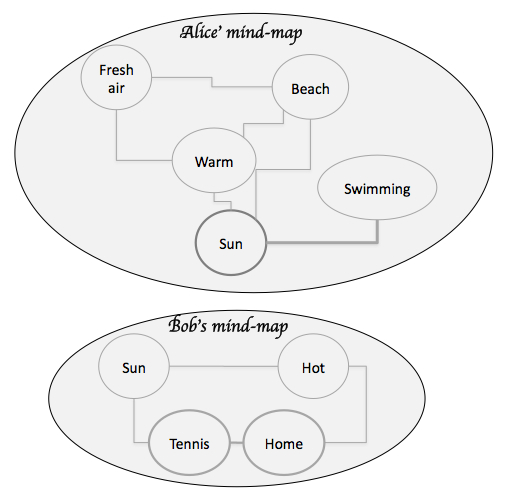} 
   \caption{\emph{Alice}'s  and \emph{Bob}'s own mind-maps}
   \label{fig:img14}
\end{figure}

Assume, that \emph{Alice}'s/\emph{Bob}'s own mind-maps look like Figure \ref{fig:img14}. Assume furthermore, that the following conversation between \emph{Bob} and \emph{Alice} takes place:
\begin{verbatim}
Bob:
The sun is shining, what a beautiful day.
Alice:
The sun is very hot.
Bob:
That is right, but I like sunny days.
\end{verbatim}

Taking into account the mind-maps of \emph{Alice} and \emph{Bob}, does \emph{Alice} now trust \emph{Bob}? Figure \ref{fig:img14} shows the different mind-maps after \emph{Bob}'s reply to \emph{Alice} - both for \emph{Alice} (what she thinks about \emph{Bob}) and \emph{Bob} (what he thinks about \emph{Alice}). The nouns and verbs are filtered out and reduced to their morphological basis. \emph{Bob}'s affirmation is inherited to \emph{Alice}'s mind-map. Please also note that the entity cell ``day'' is activated twice (because of the merge process) and that the learning principle has strengthened the connection between ``day'' and ``sunny''.

\emph{Alice}'s match value after the conversation concerning \emph{Bob} is low (0.2), because only one entity cell (``sun'') appears during the conversation with some relevance, but neither ``Fresh air'', ``warm'', ``beach'' nor ``swimming''. Concerning Bob, his match value is 1.0 because ``hot'' and ``sun'' occur during the conversation. For this, depending on each trust threshold $\alpha$, (s)he trusts or not.

 \begin{figure}[!h] 
   \centering
   \includegraphics[width=6.5cm]{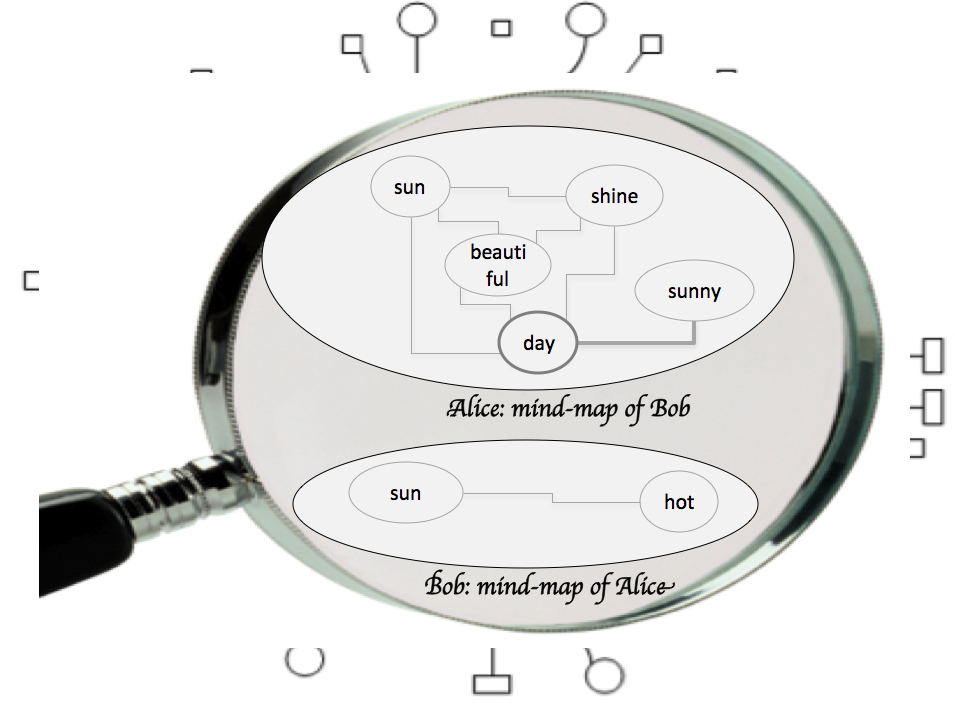} 
   \caption{\emph{Alice}'s mind map for \emph{Bob} (top) and \emph{Bob}'s mind-map for \emph{Alice} (bottom) - after the conversation.}
   \label{fig:img15}
\end{figure}

\section{Conclusions}

In the presented (position) paper, we have introduced the idea on giving explorative mind-maps a cognitive face. For this, we foster on a calculation of trust between conversing partners and argue that such an evaluation bases on the match of two existing knowledge structures. For that we have used two independent mind-maps $M_p$ (self knowledge) and $M^{*}_{pq}$ (knowledge about the conversational partner). Also the amount of trustworthiness is measured by a trust threshold $\alpha$ that is individually set. \\

Currently, we are refining the model and extending the described trust computational framework by additional substances, which include diverse experiments on matching functions, the assembling of the mind-maps per se (which probably may not depend on the conversation only), and other connectionist methods, which are not mentioned over here.

\section*{Acknowledgement}
The current work has been conducted at the University of Luxembourg. We thank the members of the MINE research group for their support.


\end{document}